\documentclass{article}

\usepackage{graphicx}

\PassOptionsToPackage{numbers, compress}{natbib}

\usepackage[final]{neurips_2024}

\usepackage[utf8]{inputenc} % allow utf-8 input
\usepackage[T1]{fontenc}    % use 8-bit T1 fonts
\usepackage{hyperref}       % hyperlinks
\usepackage{url}            % simple URL typesetting
\usepackage{booktabs}       % professional-quality tables
\usepackage{amsfonts}       % blackboard math symbols
\usepackage{nicefrac}       % compact symbols for 1/2, etc.
\usepackage{microtype}      % microtypography
\usepackage{xcolor}         % colors

\usepackage{amsmath, amsfonts}
\usepackage{graphicx}
\usepackage{xcolor}
\usepackage{wrapfig}
\usepackage{adjustbox}
\usepackage{xspace}
\usepackage{epstopdf}
\usepackage{xcolor}
\usepackage{float}
\usepackage{soul}
\usepackage{hyperref}
\usepackage{algorithm}
\usepackage{algcompatible}
\usepackage[capitalize]{cleveref}

\usepackage{eqexpl}
\eqexplSetDelim{=}

\usepackage{algpseudocode}
\usepackage{array}
\usepackage{tabularx}

\usepackage{wrapfig}
\usepackage[small]{caption}
\usepackage{subcaption}
\usepackage[noadjust]{cite}
\usepackage{comment}
\usepackage{lipsum}

\title{A Machine Learning Approach to Contact Localization in Variable Density Three-Dimensional Tactile Artificial Skin}

\makeatletter
\renewcommand\thanks[1]{#1}
\makeatother

\author{%
  \parbox[t]{\textwidth}{\centering
  Carson Kohlbrenner$^\dagger$ \ \
  Mitchell Murray$^\dagger$ \ \
  Yutong Zhang$^\dagger$ \ \
  Caleb Escobedo\\
  \vspace{5pt}
  Thomas Dunnington \ \
  Nolan Stevenson \ \
  Nikolaus Correll \ \
  Alessandro Roncone}\\
  \\
  \thanks{
    Department of Computer Science, University of Colorado Boulder
  }\\
  \thanks{$^\dagger$ represents equal contributions.}\\
  \thanks{
    \parbox[t]{\textwidth}{\centering
    \texttt{name.surname@colorado.edu}
    }
  }
}

\begin{document}

\maketitle

\begin{abstract}
Estimating the location of contact is a primary function of artificial tactile sensing apparatuses that perceive the environment through touch. Existing contact localization methods use flat geometry and uniform sensor distributions as a simplifying assumption, limiting their ability to be used on 3D surfaces with variable density sensing arrays. This paper studies contact localization on an artificial skin embedded with mutual capacitance tactile sensors, arranged non-uniformly in an unknown distribution along a semi-conical 3D geometry. 
A fully connected neural network is trained to localize the touching points on the embedded tactile sensors. The studied online model achieves a localization error of $5.7 \pm 3.0$ mm.
This research contributes a versatile tool and robust solution for contact localization that is ambiguous in shape and internal sensor distribution.
\end{abstract}

\section{Introduction}

\begin{wrapfigure}{r}{0.5\textwidth}
    \centering
    \vspace{-10pt}
    \includegraphics[width=0.45\textwidth]{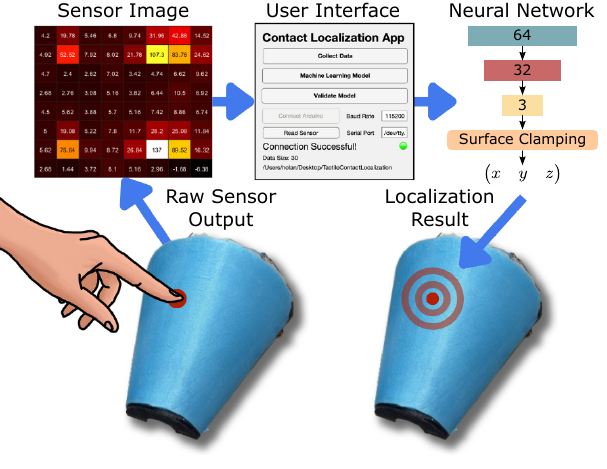}
    \caption[Cover photo caption]{Contact localization model takes in a sensor image from any configuration of artificial tactile skin and determines the location of touch through a feedforward neural network. %\protect \footnotemark.
    }
    \label{fig:cover_pic}
    \vspace{-15pt}
\end{wrapfigure}

%What is the problem and why is it important:
Artificial skin sensors are a direct way to measure external contact and have a wide range of applications in healthcare \citep{burns2020getting, erickson2019multidimensional}, prosthetics, and robotics \citep{Yang2019, cheng2019comprehensive}. As robots are increasingly deployed in close proximity to humans, the ability to localize where touch occurs allows actuating bodies to react appropriately to detected contact, interactable objects, and obstacles. 

%What are the shortcomings of current attempts at solving this problem:
Artificial skin that is capable of localizing contact to a greater number of points on its sensor array than the number of sensors present is currently limited to flat surfaces \citep{dawood2023learning, erickson2019multidimensional, wang2013mutual, zhao2022large}. The localization methods used are typically constrained by assuming either a flat plate array or a uniform density of sensors. This paper presents a method of localizing contact on 3D artificial skin with non-uniform sensor distribution. %and an added third dimension.

%What is our approach at solving this problem
We present a curved, variable-density artificial skin equipped with mutual capacitance sensors that detect touch on the surface of the artificial skin. A neural network infers a relationship between the sensor readings and the geometry of the tactile skin. We evaluate this model by comparing known touch locations to those returned by the model to characterize the accuracy and uncertainty. Our model robustly localizes the touching region on the 3D skin with variable sensor distribution. The code is available at
\url{https://github.com/HIRO-group/TactileContactLocalization}

\section{Related Work}

Artificial skin sensors have a range of sensing methods such as acoustics \citep{wall2023passive, rupavatharam2023sonicfinger}, computer vision \citep{chen2018tactile, do2022densetact, do2023densetact, dong2017improved, andrussow2023minsight}, capacitive sensing \citep{teyssier2021human, ding2019proximity, goeger2010tactile, goger2013tactile, hoshi2006robot, navarro2013methods, wang2013mutual, maiolino2013flexible}, electrical resistance tomography (ERT) \citep{park2020ert, park2022neural}, and fiber Bragg grating (FBG) optical sensing \citep{massari2022functional}. Artificial skins composed of sensor arrays have increased scalability and conformity to non-flat geometries compared to computer vision tactile sensors. Our artificial skin uses a low-cost mutual capacitance sensing array method that is highly flexible to shape and size during fabrication.

Explicit contact localization methods can achieve high accuracies with capacitive sensor arrays and are standard in touchscreens and artificial skins with known sensor distributions \citep{teyssier2021human, park2022neural}. These explicit methods utilize the known positions of sensors to interpret locations of contact. Machine learning has been used in previous artificial skins to enhance contact localization accuracy with known sensor locations due to its ability to learn complex patterns \citep{massari2022functional, iskandar2024intrinsic}. Our method treats capacitive sensor readings as images and uses a neural network to bypass the embedded spatial distribution of sensors by directly learning touch localization patterns.

\section{Method}

\subsection{Fabrication}

\begin{wrapfigure}{r}{0.5\textwidth}
    \centering
    \vspace{-10pt}
    {\includegraphics[width=0.48 \textwidth]{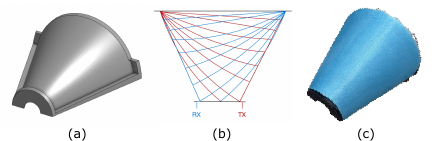} }
    \caption{(a) CAD model for the curved geometry.
    (b) General wiring scheme. Sensors are located at each intersection of transmitter (TX) and receiver (RX) wires.
    (c) Fabricated skin sensor array where the sensing circuits are embedded within a layer of silicone.}
    \label{fig:circuit}%
    \vspace{-15pt}
\end{wrapfigure}

A challenge in implementing an artificial sensing skin array on a 3D object is conforming the sensing array to a curved surface. This is addressed by fabricating a two-dimensional sheet embedded with sensors, and then overlaying the artificial skin onto a semi-conical surface.

As shown in  \cref{fig:circuit}, the curved surface is outfitted with a total of 16 wires, divided evenly into 8 transmit and 8 receive elements, forming a network of $64$ intersections. These intersections serve as the loci for capacitance measurements, effectively creating $64$ sensors distributed across the surface. A mutual capacitance board (Muca) attached to the wires measures the capacitance values at each intersection \citep{teyssier2021human}. A layer of $6mm$ thick silicone rubber is then molded on top of the wiring to provide a dielectric layer between wire intersections to ensure effective capacitive sensing. A thin copper sheet affixed to the underside of the 3D-printed structure functions as the grounding plate, mitigating electromagnetic noise from beneath the artificial skin. The result is a flush, $14.2\times16.4\times8.1$ cm semi-conical surface capable of contact localization through mutual capacitive sensing.

\subsection{Calibration}
\label{sec:calibration}
The calibration process includes collecting a small dataset in a given operational environment to establish a correlation between raw capacitance readings and specific touch locations in 3D space. Each pair of raw sensor readings corresponding to a touch location are referred to as a "point log". Data is gathered by touching the artificial skin with an index finger at known locations. Two methodologies are employed for collecting point logs: random sampling and even spacing. For random sampling, the location is chosen as a point on a randomly selected edge from the CAD model of the skin. For even spacing, the surface of the CAD model is discretized into evenly spaced points depending on the selected number of point logs. When the skin sensor array is touched, 50 samples of the capacitance readings for each sensor are stored and correlated to the prompted touch position. This process is repeated until the desired number of point logs has been processed.

\subsection{Sensing}
\label{sec:sensing}

Mutual capacitance coupling occurs at the intersection of transmitter and receiver nodes. The intensity of the coupled capacitance can be measured by sending a known signal to the transmitter wire and measuring the signal of the receiver wire. The coupled capacitance is affected by grounded conductive objects entering the electromagnetic field near the intersection, such as a human finger. Grounded conductive objects reduce the measured capacitance at the receiver electrodes which correlates to higher contact outputs for a particular sensor.

The accuracy of the sensor measurements is determined by their noise levels through a signal-to-noise ratio. Fifty sensor measurements are taken for every point log and calibration measurement. Using these measurement samples, the average sensor values for each point log ($\bar{S}$), the average initial sensor values without contact ($S_{0}$), and the standard deviation from the average sensor value without contact ($\sigma_{0}$) are used to calculate the signal-to-noise ratio (SNR) of each sensor, $i$, within the sensor array using Eq. \ref{eq:SNR}. 
\begin{equation}
    SNR_i = 20 \log_{10}\left(\frac{max(\bar{S}_i) - S_{i,0}}{\sigma_{i,0}}\right)
    \label{eq:SNR}
\end{equation}

\subsection{Contact Localization Model}

\begin{wrapfigure}{r}{0.55\textwidth}
    \centering
    \vspace{-10pt}
    \includegraphics[width=0.55\textwidth]{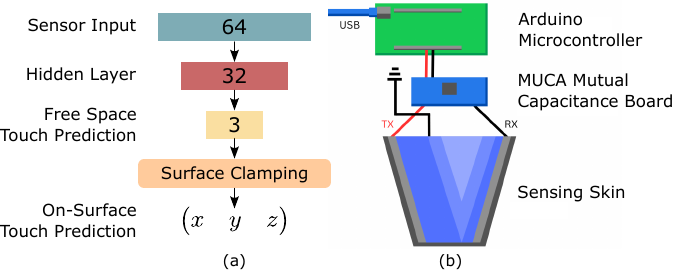}
    \caption{(a) Fully-connected neural network takes in sensor input of size $64$ and outputs a 3D coordinate.  (b) Artificial tactile skin sends mutual capacitance measurements to an Arduino microcontroller that formats the readings and passes into the neural network.}
    \label{fig:diagram}
    \vspace{-10pt}
\end{wrapfigure}

We use a supervised fully-connected neural network to localize the touching point based on raw sensor readings. This type of model is a low computation method of translation from raw sensor inputs to spatial predictions due to the nature of the artificial skin sensor data, as explained in \cref{sec:sensing}. In the ideal scenario with zero sensor noise, a sensor image exists for every possible touch position along the surface of the skin. A sensor image is the full set of sensor values for a single touch and shown in Fig \ref{fig:cover_pic}. Although this sensor image is also dependent on the probing finger and background electromagnetic field capacitance, we will assume these variables remain constant for this experiment by testing with the same finger and without changing locations for the duration of collecting data. A newly recorded sensor image can therefore be linked to a unique location along the surface of the skin once the relationship between the sensor image and position is established through our trained model.

The advantage of this method as opposed to algorithmic statistical estimation approaches is that it does not require the locations of the sensors within the skin to be known. The neural network uses a mean square error loss to compare the predicted touch locations from the sensor images collected during the calibration process against the associated given probe locations. A single hidden layer with 32 hidden nodes is trained to minimize the loss using gradient descent.
To constrain the output to the surface of the skin, the predicted output from the model is compared to a set of discrete points on the surface of the skin. The point on the surface with the shortest distance to the predicted location of the model is used as the final prediction. This process effectively constrains the model to strictly output points on the surface of the skin.

\section{Results}

\begin{figure}[pth]%{r}{0.6\textwidth}
    \centering
    \vspace{-5pt}
    {\includegraphics[width=0.99 \textwidth]{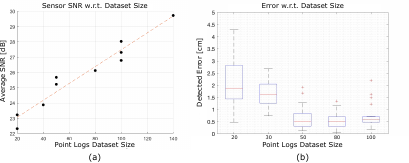} }
    \caption{(a) The linear relationship between SNR and point logs dataset size suggests correlation. (b) Prediction error for contact localization models trained with varying sets of point logs.}
    \label{fig:snr_error}%
    \vspace{-10pt}
\end{figure}

% Talk about validation error
Multiple sets of training data are collected with a varied number of point logs to analyze the accuracy of the contact localization model. \cref{fig:snr_error}(b) shows the prediction error for a validation set of 20 random point logs tested for accuracy using contact localization models trained with 20, 50, 80, and 100 point logs, respectively. Data collection took approximately 1-2 minutes for every 20 point logs. The results show a trend of increasing the number of point logs used to train the model decreases the overall error in touch detection. However, on our skin array, this decrease in error levels off at about 80 point logs. Our best model has an accuracy of $5.7 \pm 3.0 mm$. 

The average SNR of all 64 sensors in the skin array is $<30$ $dB$. \cref{fig:snr_error}(a) shows the resulting SNR values for each trained model that we have tested. There exists a linear relationship between the quantity of point logs used in training the model and SNR values. The proportional increase in SNR with calibration dataset size may be due to a wider range of activation throughout the embedded sensors during calibration.

\begin{wrapfigure}{r}{0.5\textwidth}
    \centering
    \begin{tabular}{|c|c|c|}
        \hline
         Sensor Type & Known & Acuity  \\
         & Locations & ($mm$) \\
         \hline
         
         %Acoustic \citep{fan2022enabling} & No & $20$ \\
         Human Calf \citep{corniani2020tactile} & No & $\approx$ 50.0 \\
         ERT \citep{park2022neural} & Yes & 6.6 $\pm$ 3.3 \\
         Curved Cap. (Ours) & No & 5.7 $\pm$ 3.0\\
         Human Fingertip \citep{corniani2020tactile} & No & $\approx$ 4.0 \\
         FBG \citep{massari2022functional} & Yes & $3.2 \pm 2.3$\\ 
         \textbf{Flat Cap.} \citep{teyssier2021human} & Yes & \textbf{0.5 $\pm$ 0.2} \\
         \hline
    \end{tabular}
    \caption{Our mutual capacitance sensor achieves spatial acuity consistent with sensing arrays of known distributions and human skin.}
    \label{tab:results_tab}
    \vspace{-10pt}
\end{wrapfigure}

\section{Discussion}
This paper studies a machine learning approach to contact localization on an artificial skin embedded with mutual capacitance tactile sensors, arranged non-uniformly in a semi-conical geometry. This model only requires a CAD model and touch data to distinguish the relationships between a sensor image and touch location. We demonstrate this using a complex and unmeasured internal sensor distribution. This paper demonstrates that it is unnecessary to locate the placement of internal sensors in artificial skin to acquire accurate touch predictions. Changes in sensor location due to skin deformation that is incurred through conformation to various surface geometries may be navigated through neural network adaptability rather than altering the fabrication and calibration design. Our method achieved a better average localization accuracy than human skin and comparable results to artificial sensors with non-uniform sensor distribution methods.

One of the biggest limitations of our current implementation is the lack of precise visual cues during the data collection process, resulting in probing inaccuracies apart from the model inaccuracies. To improve this, we propose the application of a grid pattern to the surface of the skin with discrete touch locations. This grid can be loaded into the contact localization model and integrated into the data collection process reducing the error between the intended and actual touch locations. In addition, this research has only been conducted using single-touch interactions. Future work aims at identifying multi-touch accuracy and gesture identification, such as swiping up or down, which will drive its application toward robot-human communication through touch.

\bibliographystyle{plainnat}
\bibliography{references}

\end{document}